\documentclass[runningheads]{llncs}
\usepackage{graphicx}
\usepackage{subfig}
\usepackage{changepage}
\usepackage[table,xcdraw]{xcolor}
\usepackage{array}
\usepackage{multirow}
\begin{document}
\title{All You Need Is Sex for Diversity}
%
%

\author{José Maria Simões\orcidID{0000-0002-6916-3217} \and
Nuno Lourenço\orcidID{0000-0002-2154-0642} \and
Penousal Machado\orcidID{0000-0002-6308-6484}}

\authorrunning{J. M. Simões et al.}

\institute{CISUC, Department of Informatics Engineering, University of Coimbra\\
Polo II - Pinhal de Marrocos, 3030 Coimbra, Portugal \\
\email{josecs@student.dei.uc.pt and \{naml,machado\}@dei.uc.pt}}

\maketitle              
\begin{abstract}
Maintaining genetic diversity as a means to avoid premature convergence is critical in Genetic Programming. Several approaches have been proposed to achieve this, with some focusing on the mating phase from coupling dissimilar solutions to some form of self-adaptive selection mechanism. In nature, genetic diversity can be the consequence of many different factors, but when considering reproduction Sexual Selection can have an impact on promoting variety within a species. Specifically, Mate Choice often results in different selective pressures between sexes, which in turn may trigger evolutionary differences among them. Although some mechanisms of Sexual Selection have been applied to Genetic Programming in the past, the literature is scarce when it comes to mate choice. Recently, a way of modelling mating preferences by ideal mate representations was proposed, achieving good results when compared to a standard approach. These mating preferences evolve freely in a self-adaptive fashion, creating an evolutionary driving force of its own alongside fitness pressure. The inner mechanisms of this approach operate from personal choice, as each individual has its own representation of a perfect mate which affects the mate to be selected. \par
In this paper, we compare this method against a random mate choice to assess whether there are advantages in evolving personal preferences.
We conducted experiments using three symbolic regression problems and different mutation rates. The results show that self-adaptive mating preferences are able to create a more diverse set of solutions when compared to the traditional approach and a random mate approach (with statistically significant differences) and have a higher success rate in three of the six instances tested.
 \par

\end{abstract}
\section{Introduction}
Natural Selection -- or survival of the fittest, as described by Charles Darwin ~\cite{Darwin_Origins} -- operates as the foundation for Genetic Programming (GP) as it does for other classes of Evolutionary Algorithms (EA). The adaptation of basic evolutionary principles has made nature-inspired algorithms suitable for several optimization tasks, as encoded solutions in a population evolve under selective pressure to solve specific problems ~\cite{Poli_Field_Guide,Eiben2015,Back_1997}. By combining solutions into new ones (crossover) and/or changing existing ones (mutation), we aim at improving results in a given area (exploitation) while promoting a broad search for new regions (exploration) ~\cite{Eiben2015_Book,ConvergenceEiben98}. \par
One of the main issues regarding EAs is that of premature convergence, where the population loses general diversity by focusing on one single promising area, which eventually leads to an inescapable set of similar solutions ~\cite{Eiben2015_Book}. As such, the balance between exploration and exploitation is important in the search for the global optima, meaning that maintaining diversity throughout the evolutionary process is paramount ~\cite{Poli_Field_Guide,Eiben2015_Book,Diversity_in_GP_1,Diversity_Measures}. \par

Drawing inspiration from natural phenomena, some authors have studied the potential benefits of Sexual Selection as a means for diversity maintenance. Firstly proposed by Darwin as a way to support the idea that Natural Selection alone could not justify certain traits that seemed to hinder survival, Sexual Selection has since then evolved to be recognized as an important  (and often complex) evolutionary force ~\cite{gayon2010,Alonzo2019,Brock2007,Ralls2009}. Whether acting on the same sex (e.g., battles for mates) or on the opposite sex (e.g., expressing preferences for certain attributes), Sexual Selection can promote the development of different and diverse traits ~\cite{Jones2009}. \par

Despite the seeming potential of Sexual Selection mechanisms in EAs, literature on this evolutionary force is scarce when it comes to its application in GP. Although not extensively, studies that incorporate Sexual Selection mechanisms in Genetic Algorithms are more common, where we can find gender separation ~\cite{Drezner2006,Bandyopadhyay1998,Zhu2006}, multiple genders ~\cite{Vrajitoru2008} or even dissimilar Mate Choice ~\cite{Varnamkhasti2012,Jalali2012}, to name a few. 
Mate Choice based on dissimilarity has also been studied in GP ~\cite{Fry2005} as well as self-adapting mate selection functions ~\cite{Tauritz2007}, yet the array of works in the field appears to become even narrower when we consider mainly individual preferences as a means to avoid early convergence. Recently, Leitão ~\cite{Leitao_Teste} proposed the PIMP method: Ideal partner representations as a way of modelling mate choice in GP.  In PIMP, each individual has two chromosomes: one that represents the solution to the problem being tackled, and another that encodes the ideal solution (or partner) against which potential mates are compared. The proposed framework pointed towards performance improvements in several symbolic regression experiments when compared to a standard approach. Furthermore, the author presents a thorough study of the resultant dynamics, which also seems to be promoting exploratory gains. Throughout the reported analysis, some other interesting factors are worth noting, such as the emergence of different roles (i.e., female (or chooser) and male (courter)) merely resulting from distinct selective pressures. \par

Given the intricate nature of Sexual Selection itself (especially Mate Choice) and the potential benefits of transposing such mechanisms to GP, we set out to expand the existing literature on the topic by shedding light on what we consider to be a relevant question: Are there benefits in encoding ideal mates as opposed to a random mate choice?
For that, we took advantage of the good results provided by the PIMP method and tested it against a simpler method in which the courter is chosen at random. In this work, a Standard Approach is also included for reference, and results are measured through performance and diversity metrics. Our results suggest that Mate Choice via the PIMP method is able to promote and maintain more diversity without compromising performance. Furthermore, the experiments suggest that the dynamics behind this mate choice mechanism behave differently from a simpler random mate choice.
\par 
The remaining of the article is organized  as follows: We first describe the PIMP approach as presented by Leitão ~\cite{Leitao_Teste} and proceed to clarify our motivation behind this experiment. We then describe the methodology (such as the metrics used to validate the comparisons) and present the results followed by an analysis and discussion of the findings.

\section{The PIMP Approach}
Proposed by Leitão ~\cite{Leitao_Teste}, PIMP (Mating Preferences as Ideal Mating Partners in the Phenotype Space) incorporates Mate Choice in GP by encoding a representation of an ideal partner. This approach is built upon a phenomenon often observed in nature regarding the preliminary stages of reproduction: a candidate (courter) has to get access to the opposite sex (chooser), which can set different selection pressures on each ~\cite{Jones2009}. Of the mechanisms within Sexual Selection, this approach models preferences, meaning that mates are chosen based on attractiveness regardless of their fitness. When applied to GP, these different pressures act in parallel and independently, having the potential of driving the evolutionary process in different directions. These dynamics are intended to avoid early convergence in a self-adaptive fashion, as preferences evolve unpredictably, whereas fitness pressure is established at the beginning of the simulation.

\par
Essentially, PIMP differs from a Standard Approach at two levels: the individuals and the selection phase, as described below.\par

\begin{figure}[b]
  \centering
  \includegraphics[width=2.3in]{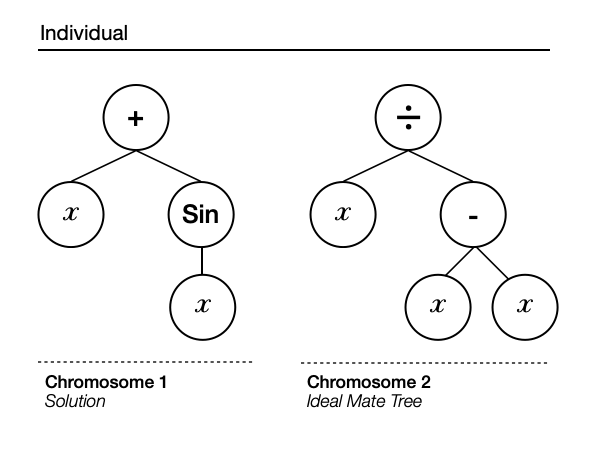}
  \caption{Illustration of the composition of an individual under the PIMP approach}
  \label{imagem_representation}
\end{figure}

\vspace{3mm}
\textbf{Individuals} \par
Each individual has two chromosomes: the solution and the encoded preference (see Figure \ref{imagem_representation}). Preferences are representations of the ideal mates, where the trees can be structurally identical to solutions, created from the same building blocks. This means that these can be generated from the same set of terminals and functions used to solve the problem at hand. The preference chromosome has no direct impact on fitness, being computed only during selection.  \par

\begin{figure}[ht]
  \centering
  \includegraphics[width=3.5in]{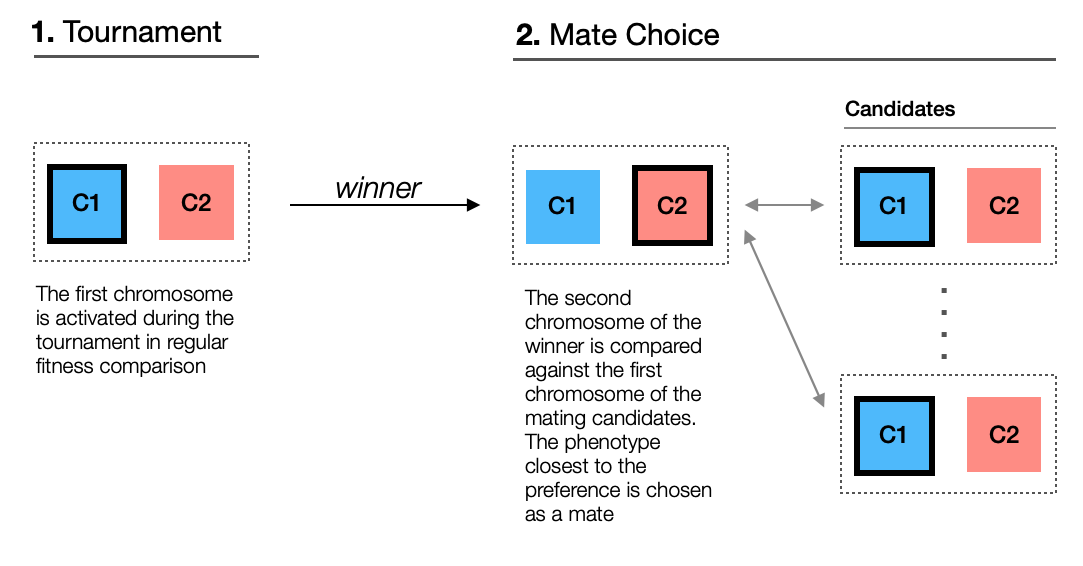}
  \caption{Illustration of the selection scheme in PIMP}
  \label{selection}
\end{figure}

\vspace{3mm}
\textbf{Selection} \par
During selection, an individual is chosen via a standard tournament, becoming the chooser. The candidates (i.e., potential partners) are randomly selected from the population. Each candidate is then compared to the second chromosome of the chooser, which is done in the same way as comparing a solution to the fitness function. In a way, one could say that the encoded preference of the chooser becomes the fitness function for that set of candidates. After going through all candidates, the one closest to the preference is chosen. A simple illustration of the selection phase is presented in Figure \ref{selection}. Finally, it is important to note that both chromosomes are subject to recombination and mutation.  \par

\section{Motivation}
In order to establish a comparison between random mate selection and Mate Choice, we decided to implement the aforementioned PIMP method by Leitão ~\cite{Leitao_Teste}. This choice was based on two major factors. First, the author provided an extensive analysis of the method and specifically targeted Mate Choice as the primary research subject, which is the main focal point of this article. Secondly, the method needs few implementation adjustments compared to a Standard Approach as encoded preferences are built upon the same arguments used to build the solutions. \par
Furthermore, the work suggests that modelling Mate Choice as in PIMP is advantageous in GP. The method showed performance gains when compared to a Standard Approach on 52 symbolic regression instances, reaching statistical significance on more than half when mutation was introduced. By having mates being selected through ideal representations, the potential mates are subject to a secondary force besides Natural Selection (via fitness function) which led to a role separation that evolved organically. As explained by the author, the self-adaptive nature of mating preferences coupled with Natural Selection promotes a more exploratory search, which in turn seems to be related to the observed performance gains. Also, it was stated that preferences do not always evolve in the same direction as the fitness function, meaning that in some test-cases divergence was observable and evolved independently of fitness pressure. \par
Hence, we hypothesize that testing this mechanism against a random mate choice might help us understand even better to what extent this approach is favourable. Under the PIMP framework, only the first parent is subject to fitness pressure, while the mate is subject to a direct comparison with the ideal solution encoded in the chooser (i.e., the first parent). By avoiding assembling couples merely through fitness metrics we allow more diversity among the population, which in turn improves performance. As such, we aim at exploring to what extent can a random mate choice lead to similar outcomes. \par
Moreover, considering that there is strong evidence that in PIMP preferences evolve in a self-reinforced fashion rather than arbitrarily -- thus having the potential of evolving towards its own goal --, we will also be able to test this against a mate choice with no sustainable direction and access potential benefits.
In the next section, we address the criteria used for this experiment.  \par

\section{Methodology}
This study compares two existing approaches to an experimental one: PIMP, Random Mate Choice (first parent chosen via tournament selection and a mate chosen at random) and a Standard Approach acting as a reference where both parents are chosen through tournaments (see Table \ref{Approaches}). The setup used for all these methods is shown in Table \ref{Set_up}. Two different mutation probabilities are considered (of 5\% and 10\%). Regarding the genetic operators, Subtree mutation is applied and One Point Crossover is used for recombination. Three symbolic regression instances were used: Koza-1 (one variable), Nguyen-6 (one variable) and Pagie-1 (two variables) (a summary of these instances can be found in~\cite{GP_Bench}). These functions were tested in the original work, thus providing us with some knowledge beforehand, particularly regarding Mean Best Fitness. While tested under Koza-1 (with a mutation rate of 1\%) PIMP showed no statistically significant differences when compared to a Standard Approach, under Nguyen-6 PIMP performed better with significant differences. We also included Pagie-1 (where PIMP also performed better with significant results) given that it can be more difficult to solve than the last two ~\cite{GP_Bench}. \par

All testing instances were performed in sets of 30 runs, and each run had its exclusive seed shared with all three approaches. Although the main setup is similar to the original work on PIMP, some changes were performed, such as the number of generations (500 originally) and the mutation rates (1\% in the original work). By increasing the number of generations, we can have a better understanding of how the population behaves in the long run, while by having higher mutation rates we simultaneously promote more genetic diversity (especially in the Standard Approach). \par

\begin{table}[ht]
\caption{Approaches used}\label{Approaches}
\addtolength\tabcolsep{10pt} 
\renewcommand\arraystretch{1.4}

\begin{adjustwidth}{1cm}{} 
\begin{tabular}{lll}
\hline
\multirow {2}{4em}{PIMP}        & Parent 1 (Chooser) & Tournament Selection (size = 5)                                       \\
                             & Parent 2 (Courter) & Random Set (size = 5) \\ \hline
\multirow {2}{4em}{Random Mate} & Parent 1           & Tournament Selection (size = 5)                                       \\
                             & Parent 2           & Random                                                                \\ \hline
Standard                     & Both Parents       & Tournament Selection (size = 5)                                       \\ \hline
\end{tabular}
\end{adjustwidth}
\end{table}


\begin{table}[t]

\caption{General Set up}\label{Set_up}
\addtolength\tabcolsep{10pt} 
\renewcommand\arraystretch{1.4}

\begin{adjustwidth}{1cm}{} 
\begin{tabular}{lllll}
\cline{1-3}
\multirow {3}{0pt} & Population Size      & 100             &  &  \\
            {General Parameters}                        & Generations          & 1500            &  &  \\
                                    & Elitism              & None            &  &  \\ \cline{1-3}
Individual Builder                  & Ramped half-and-half & random(2,6)     &  &  \\ \cline{1-3}
\multirow  {3}{0pt}           & Crossover Prob.      & 0.9             &  &  \\
           {Breeding}                         & Mutation Prob.       & {[}0.05, 0.1{]} &  &  \\
                                    & Max Depth            & 17              &  &  \\ \cline{1-3}
\end{tabular}
\end{adjustwidth}
\end{table}

\subsection{Measures}

We compare the mentioned approaches at two main levels: performance and diversity. The metrics used for each are described below. \par
\vspace{3mm}
\textbf{Performance Measures} \par
To assess performance differences we compute the Mean Best Fitness (MBF) measured through Mean Squared Error for all algorithms under study. Considering that we are testing them under symbolic regression instances with known solutions, we defined a success rate regarding the quality of the solutions. Therefore, we have: \par

\begin{itemize}
  \item Mean Best Fitness (Best individual of the final population in each run)
  \item Success Rate (Number of runs (out of 30) in which at least one solution is better than 1E-4)
\end{itemize}

\textbf{Diversity Measures} \par
While performance metrics are quite straightforward to establish, diversity metrics require a bit more discussion. When studying diversity, different techniques can be applied to examine its dynamics and effects within the population ~\cite{Diversity_in_GP_1}. Knowing that in GP different trees can have the same fitness, we decided not to measure population diversity by fitness variety. Instead, we performed a comparison at the level of the genotype to identify differences between trees (e.g., through Tree Edit Distance ~\cite{Sasha_TreeEdit}). Nevertheless, this choice came with a cost that proved to be too expensive computation-wise. As such, it was decided to record only the number of unique solutions in the population every one hundred generations. This metric is not sufficient to tell us how different solutions are, but it can nonetheless be useful at a higher level: if an algorithm is not able to maintain unique solutions, then convergence is more likely to occur. \par
Another measure used to assess diversity is strictly linked to the original work on PIMP. In the experiments, the author opted for operation restrictions at the root node -- which was never chosen as a crossover point or for mutation. This means that if these particular alleles managed to avoid extinction, they were forced to remain the same throughout the evolutionary process. Furthermore, under these conditions, if a particular root node is lost it is then lost forever. Thus, it was decided to keep this particular feature in implementation as we saw it as a way of studying which approach was better at maintaining diversity at the root node. To achieve this, we kept track of each individual's root node throughout evolution. This was a straightforward process as all functions used for testing share the same function set (see Table~\ref{table_function_set}). A run where there the root node is the same in every individual on the last generation was considered a convergence instance. With this, we can list the two metrics used for diversity analysis:\par

\begin{itemize}
    \item Number of unique solutions among the population
    \item Runs where root node convergence was avoided
\end{itemize}

\begin{table}

\caption{Function Set (same for all instances)}\label{table_function_set}
\addtolength\tabcolsep{10pt} 
\renewcommand\arraystretch{1.4}
\begin{adjustwidth}{2.5cm}{} 
\begin{tabular}{lllll}
\multicolumn{1}{c}{Functions} & \multicolumn{1}{c}{Constants} &  &  &  \\ \cline{1-2}
\multicolumn{1}{c}{ $ +, -, \times, \%, \sin, \cos, e^n, \ln(|n|) $}        & \multicolumn{1}{c}{None}      &  &  &  \\
                              &                               &  &  &  \\
                              &                               &  &  & 
\end{tabular}
\end{adjustwidth}
\end{table}

\subsection{Statistical Tests}
After running the experiments and gathering the required data for analysis, a set of statistical tests was performed to better understand the significance of the observed differences. We highlight that in this experiment we are dealing with dependent samples, given that each run had a specific initial population.\par
For tests consisting of quantitative data (Mean Best Fitness (MBF) and Number of Unique Solutions (NUS)), a Shapiro-Wilk test was performed to assess the likelihood of having normally distributed data, while Bartlett's test was performed to test the variance of the data. For an alpha value of 5\%, MBF data failed the normality assumption, and while NUS data only failed to meet the normality assumption under the function Pagie-1, the Bartlett's test showed no signs of equal variance between groups (also for an alpha of 5\%). As such, a Friedman test was performed to test the differences between groups. Whenever differences were found, a Wilcoxon signed rank was performed as a post hoc with the correspondent Bonferroni correction \cite{stats_book}.\par
For qualitative data (Success Rate and Root Node convergence), a Cochran's Q test was used to find significant differences between approaches, proceeding with a McNemar's Test whenever that was the case. Again, an alpha value of 5\% was used when comparing the three approaches, and a Bonferroni correction was applied for paired comparisons \cite{stats_book}.

\section{Results}

In this section we present the results for each symbolic regression instance and mutation variation. For the sake of readability, Mean Best Fitness results are presented in two different tables (one for each mutation variant) as the sample standard deviation is also included. Although the raw data regarding the success rates is nominal, results are presented in the percentage form merely on a stylistic choice. Whenever there is statistical significance between PIMP and Random Mate approach these will be highlighted (\textbf{bold}). Furthermore, an asterisk sign (\textbf{*}) next to a result indicates that that particular approach is statistically different from the Standard Approach. \par

\vspace{0.5cm}
\textbf{Mean Best Fitness}\par
Table \ref{mbf_5} and Table \ref{mbf_10} depict the results regarding the MBF for different mutation rates as well as the sample standard deviation. From a general overview, PIMP does not always outperform the Random Mate approach, which in turn shows slightly better results in some instances. Similarly, whenever PIMP shows better results, the differences between these two approaches do not seem to be critical. Furthermore, including the Standard Approach as a benchmark, PIMP seems to do better also with a seemingly small margin. Using a Random Mate approach seems to provide competitive results to a Standard one, which might suggest that under these specific conditions having both parents under fitness pressure is not necessarily advantageous. This might be reinforced by the fact that no statistically significant differences were found. However, it must be mentioned that on the particular instance of the function Pagie-1 with a mutation rate of 10\%, we obtained a \textit{p-value} of $ \approx 0.048 $. Although this meant that differences were significant under the defined criteria, this \textit{p-value} was considered to be too close to the chosen alpha value. This was later confirmed when performing a Wilcoxon signed rank, where no differences were found. As such, we have decided to follow a conservative approach and conclude that no relevant differences were found.\par


\begin{table}[ht]
\caption{Results for the MBF and StDev with a mutation rate of 5\% over 30 runs}\label{mbf_5}
\addtolength\tabcolsep{10pt} 
\renewcommand\arraystretch{1.4}

\begin{adjustwidth}{0.7cm}{} 
\begin{tabular}{llccc}
                                                &                                                    & \multicolumn{1}{l}{PIMP}                               & \multicolumn{1}{l}{Random Mate}                        & \multicolumn{1}{l}{Standard}                           \\ \hline
\multicolumn{1}{c|}{}                           & \cellcolor[HTML]{EFEFEF}MBF                        & \cellcolor[HTML]{EFEFEF}3.44E-4                        & \cellcolor[HTML]{EFEFEF}1.75E-3                        & \cellcolor[HTML]{EFEFEF}9.68E-4                        \\ \cline{2-5} 
\multicolumn{1}{c|}{\multirow{-1.9}{4em}{Koza-1}}   & StDev                                              & 4.86E-4                                                & 6.29E-3                                                & 2.20E-3                                                \\ \hline
\multicolumn{1}{r|}{}                           & \cellcolor[HTML]{EFEFEF}{\color[HTML]{000000} MBF} & \cellcolor[HTML]{EFEFEF}{\color[HTML]{000000} 6.54E-3} & \cellcolor[HTML]{EFEFEF}{\color[HTML]{000000} 1.00E-2} & \cellcolor[HTML]{EFEFEF}{\color[HTML]{000000} 5.82E-3} \\ \cline{2-5} 
\multicolumn{1}{r|}{\multirow{-1.9}{5em}{Nguyen-6}} & StDev                                              & 1.72E-2                                                & 2.33E-2                                                & 1.52E-2                                                \\ \hline
\multicolumn{1}{l|}{}                           & \cellcolor[HTML]{EFEFEF}MBF                        & \cellcolor[HTML]{EFEFEF}9.11E-3                        & \cellcolor[HTML]{EFEFEF}1.29E-2                        & \cellcolor[HTML]{EFEFEF}1.57E-2                        \\ \cline{2-5} 
\multicolumn{1}{l|}{\multirow{-1.9}{4em}{Pagie-1}}  & StDev                                              & 1.06E-2                                                & 1.23E-2                                                & 1.46E-2                                                \\ \hline
\end{tabular}
\end{adjustwidth}
\end{table}



\begin{table}
\caption{Results for the MBF and StDev with a mutation rate of 10\% over 30 runs}\label{mbf_10}
\addtolength\tabcolsep{10pt} 
\renewcommand\arraystretch{1.4}

\begin{adjustwidth}{0.7cm}{} 
\begin{tabular}{llccc}
                                                &                                                    & \multicolumn{1}{l}{PIMP}                               & \multicolumn{1}{l}{Random Mate}                        & \multicolumn{1}{l}{Standard}                           \\ \hline
\multicolumn{1}{c|}{}                           & \cellcolor[HTML]{EFEFEF}MBF                        & \cellcolor[HTML]{EFEFEF}5.56E-4                        & \cellcolor[HTML]{EFEFEF}1.28E-3                        & \cellcolor[HTML]{EFEFEF}3.9E-4                         \\ \cline{2-5} 
\multicolumn{1}{c|}{\multirow{-1.9}{4em}{Koza-1}}   & StDev                                              & 1.01E-3                                                & 4.75E-3                                                & 8.59E-4                                                \\ \hline
\multicolumn{1}{r|}{}                           & \cellcolor[HTML]{EFEFEF}{\color[HTML]{000000} MBF} & \cellcolor[HTML]{EFEFEF}{\color[HTML]{000000} 5.63E-3} & \cellcolor[HTML]{EFEFEF}{\color[HTML]{000000} 2.74E-3} & \cellcolor[HTML]{EFEFEF}{\color[HTML]{000000} 6.23E-3} \\ \cline{2-5} 
\multicolumn{1}{r|}{\multirow{-1.9}{5em}{Nguyen-6}} & StDev                                              & 1.5E-2                                                 & 8.57E-3                                                & 1.76E-2                                                \\ \hline
\multicolumn{1}{l|}{}                           & \cellcolor[HTML]{EFEFEF}MBF                        & \cellcolor[HTML]{EFEFEF}9.62E-3                        & \cellcolor[HTML]{EFEFEF}1.08E-2                        & \cellcolor[HTML]{EFEFEF}1.50E-2                        \\ \cline{2-5} 
\multicolumn{1}{l|}{\multirow{-1.9}{4em}{Pagie-1}}  & StDev                                              & 1.08E-2                                                & 1.21E-2                                                & 1.44E-2                                                \\ \hline
\end{tabular}
\end{adjustwidth}
\end{table}

\vspace{3mm}
\textbf{Success Rate}\par
Just as observed in the MBF results, the data presented in Table \ref{success_rate} also does not seem to provide any consistent advantage for any approach when it comes to success rate. The main differences can be found on Nguyen-6 with a mutation rate of 5\%, where PIMP  and Random Mate promote a gain of 26.7\% and 23.3\%, respectively, over the Standard Approach. Conversely, by increasing the mutation rate to 10\% in the same function the Standard Approach performs better, with a success rate 26.7\% better than PIMP and only 3.4\% over a Random Mate choice. On both Pagie-1 instances (which is considered to be more difficult to tackle \cite{GP_Bench}), a Random Mate approach fails to find any good solution under the defined terms, while PIMP manages to do so in 2 runs out of 30. Performing a  Cochran's Q test on the data results in no significance for any instance.\par


\begin{table}[ht]
\caption{Success Rate (average of 30 runs)}\label{success_rate}
\addtolength\tabcolsep{10pt} 
\renewcommand\arraystretch{1.4}

\begin{adjustwidth}{0.6cm}{} 
\begin{tabular}{lcccc}
                                                & \multicolumn{1}{l}{Mutation \%}                    & \multicolumn{1}{l}{PIMP}                            & \multicolumn{1}{l}{Random Mate}                       & \multicolumn{1}{l}{Standard}                          \\ \hline
\multicolumn{1}{c|}{}                           & \cellcolor[HTML]{EFEFEF}5\%                        & \cellcolor[HTML]{EFEFEF}56.6\%                      & \cellcolor[HTML]{EFEFEF}56.6\%                        & \cellcolor[HTML]{EFEFEF}60\%                          \\ \cline{2-5} 
\multicolumn{1}{c|}{\multirow{-1.9}{4em}{Koza-1}}   & 10\%                                               & 60\%                                                & 53.3\%                                                & 43.3\%                                                \\ \hline
\multicolumn{1}{r|}{}                           & \cellcolor[HTML]{EFEFEF}{\color[HTML]{000000} 5\%} & \cellcolor[HTML]{EFEFEF}{\color[HTML]{000000} 60\%} & \cellcolor[HTML]{EFEFEF}{\color[HTML]{000000} 56.6\%} & \cellcolor[HTML]{EFEFEF}{\color[HTML]{000000} 33.3\%} \\ \cline{2-5} 
\multicolumn{1}{r|}{\multirow{-1.9}{5em}{Nguyen-6}} & 10\%                                               & 43.3\%                                              & 66.6\%                                                & 70\%                                                  \\ \hline
\multicolumn{1}{l|}{}                           & \cellcolor[HTML]{EFEFEF}5\%                        & \cellcolor[HTML]{EFEFEF}6.6\%                       & \cellcolor[HTML]{EFEFEF}0\%                           & \cellcolor[HTML]{EFEFEF}3.3\%                         \\ \cline{2-5} 
\multicolumn{1}{l|}{\multirow{-1.9}{4em}{Pagie-1}}  & 10\%                                               & 0\%                                                 & 0\%                                                   & 3.3\%                                                 \\ \hline
\end{tabular}
\end{adjustwidth}
\end{table}


\vspace{0.5cm}
\textbf{Diversity Analysis}\par
\begin{figure}[ht]
  \centering
  \subfloat[Koza-1 Mutation rate 5\%]{\includegraphics[width=0.7\textwidth]{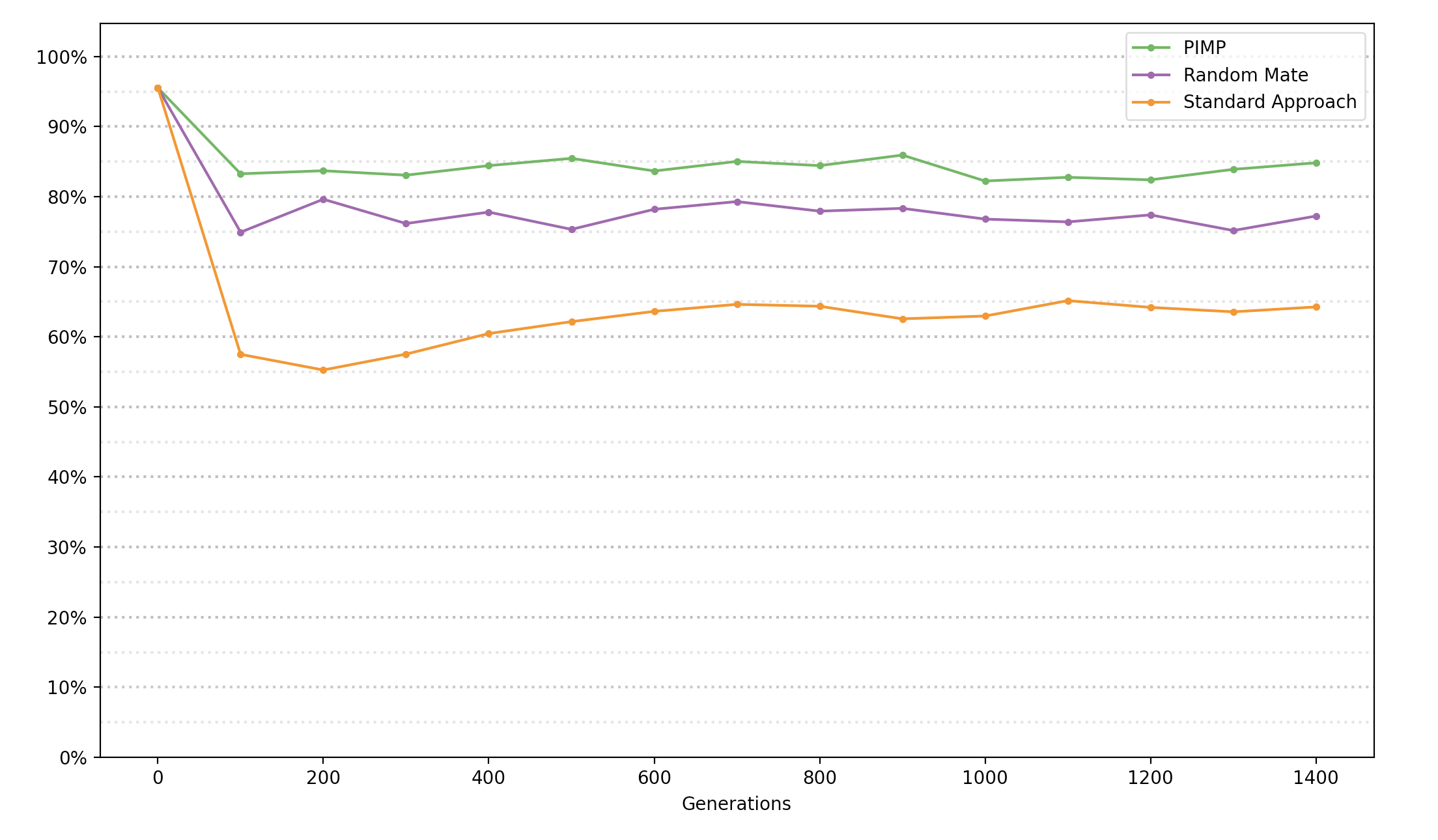}\label{imagem_unique1}}
  \hfill
  \subfloat[Pagie-1 Mutation rate 5\%]{\includegraphics[width=0.7\textwidth]{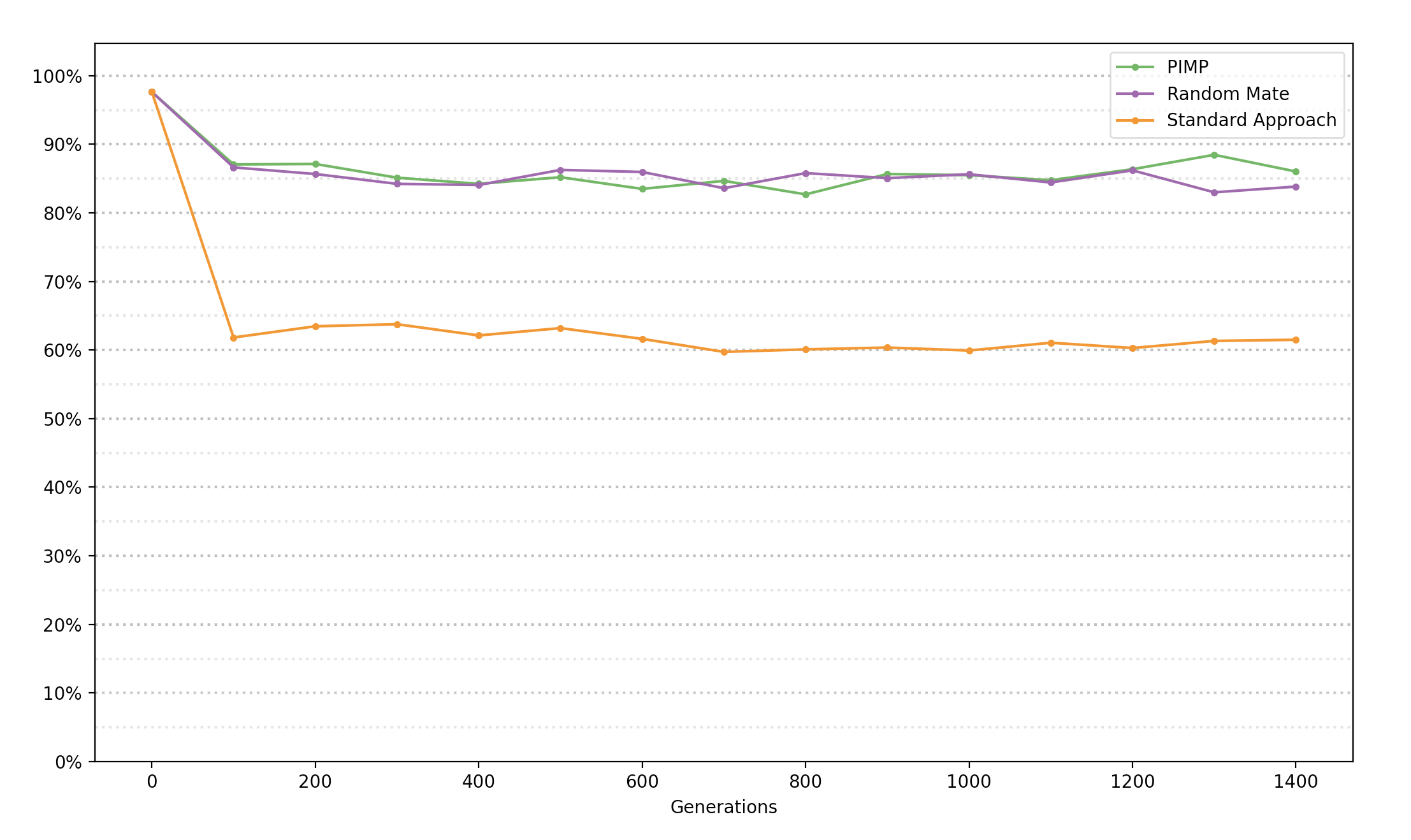}\label{imagem_unique2}}
  \caption{Percentage of unique solutions each 100 generations (average).}
\end{figure}

While there seems to be no performance gains in using any of the studied approaches over the others, the same can not be said regarding diversity maintenance. Table \ref{unique_sol} depicts the results on the average percentage of unique trees after 1500 generations, where there is a consistent pattern throughout all instances. It can be observed that under the Standard Approach the final population tended to contain roughly 35\% of repeated solutions on average, which might be the consequence of submitting both parents to fitness pressure. In terms of genetic diversity, Random Mate provides an improvement over this, which is further improved when using PIMP, where on average it produced 15\% of repeated solutions. Statistic results from Friedman tests show significant differences in all instances, which was later confirmed through a paired Wilcoxon Signed Rank holding two exceptions: in Pagie-1 with mutation rates of 5\% and 10\%, PIMP and Random Mate have no significant differences between them. Figures \ref{imagem_unique1} and \ref{imagem_unique2} illustrate the percentage of unique solutions throughout the evolutionary process each 100 generations for Koza-1 (mutation rate of 5\%) and for Pagie-1 (mutation rate of 5\%)\par


\begin{table}[ht]
\caption{Unique Solutions After 1500 Generations (Average of 30 runs)}\label{unique_sol}
\addtolength\tabcolsep{10pt} 
\renewcommand\arraystretch{1.4}

\begin{adjustwidth}{0.5cm}{} 
\begin{tabular}{lcccc}
                                                & \multicolumn{1}{l}{Mutation \%}                    & \multicolumn{1}{l}{PIMP}                              & \multicolumn{1}{l}{Random Mate}                     & \multicolumn{1}{l}{Standard}                          \\ \hline
\multicolumn{1}{c|}{}                           & \cellcolor[HTML]{EFEFEF}5\%                        & \cellcolor[HTML]{EFEFEF}\textbf{84.8}\%*                        & \cellcolor[HTML]{EFEFEF}\textbf{77.2}\%*                      & \cellcolor[HTML]{EFEFEF}64.26\%                       \\ \cline{2-5} 
\multicolumn{1}{c|}{\multirow{-1.9}{4em}{Koza-1}}   & 10\%                                               & \textbf{86.4}\%*                                                & \textbf{77.4}\%*                                              & 66.6\%                                                \\ \hline
\multicolumn{1}{r|}{}                           & \cellcolor[HTML]{EFEFEF}{\color[HTML]{000000} 5\%} & \cellcolor[HTML]{EFEFEF}{\color[HTML]{000000} \textbf{84.2}\%*} & \cellcolor[HTML]{EFEFEF}{\color[HTML]{000000} \textbf{79}\%*} & \cellcolor[HTML]{EFEFEF}{\color[HTML]{000000} 63.2\%} \\ \cline{2-5} 
\multicolumn{1}{r|}{\multirow{-1.9}{5em}{Nguyen-6}} & 10\%                                               & \textbf{86.8}\%*                                                & \textbf{78.3}\%*                                              & 62.2\%                                                \\ \hline
\multicolumn{1}{l|}{}                           & \cellcolor[HTML]{EFEFEF}5\%                        & \cellcolor[HTML]{EFEFEF}86\%*                          & \cellcolor[HTML]{EFEFEF}83.8\%*                      & \cellcolor[HTML]{EFEFEF}61.5\%                        \\ \cline{2-5} 
\multicolumn{1}{l|}{\multirow{-1.9}{4em}{Pagie-1}}  & 10\%                                               & 88.4\%*                                                & 84.6\%*                                              & 67.6\%                                                \\ \hline
\end{tabular}
\end{adjustwidth}
\end{table}


\vspace{5mm}
\textbf{Avoiding Root Node Convergence} \par
Finally, Table \ref{root_conv} provides an overview of the ability of each approach to prevent convergence to a single function at the root node. Similarly, there is a gradual improvement from the Standard Approach to Random Mate, followed by PIMP. These results are also an interesting aspect of this study considering that, as discussed earlier, preserving root node diversity under the current conditions is a difficult task. Although there seems to be an advantage in using PIMP to achieve this (mainly in Koza-1 and Nguyen-6), the only instances where statistical significance was reached was in Koza-1 with a mutation rate of 5\%, where PIMP was statistically different from Random Mate and Standard Approach (\textit{p-values} $ \approx 0.004 $ and $ \approx 0.0009 $ respectively) and in Pagie-1 with a mutation rate of 10\% -- both PIMP and Random Mate held no differences (\textit{p-value} $ \approx 0.54 $) but both were statistically different from the Standard Approach (\textit{p-values} of $ \approx 0.0009 $ and $ \approx 0.002 $). \par


\begin{table}[hb!]
\caption{Instances where root node convergence was avoided after 1500 Generations (Out of 30 runs)}\label{root_conv}
\addtolength\tabcolsep{10pt} 
\renewcommand\arraystretch{1.4}

\begin{adjustwidth}{0.3cm}{} 
\begin{tabular}{lcccc}
                                                & \multicolumn{1}{l}{Mutation \%}                    & \multicolumn{1}{l}{PIMP}                            & \multicolumn{1}{l}{Random Mate}                     & \multicolumn{1}{l}{Standard}                        \\ \hline
\multicolumn{1}{c|}{}                           & \cellcolor[HTML]{EFEFEF}5\%                        & \cellcolor[HTML]{EFEFEF}\textbf{11/30}*                       & \cellcolor[HTML]{EFEFEF}\textbf{3/30}                        & \cellcolor[HTML]{EFEFEF}0/30                        \\ \cline{2-5} 
\multicolumn{1}{c|}{\multirow{-1.9}{4em}{Koza-1}}   & 10\%                                               & 12/30                                               & 7/30                                                & 5/30                                                \\ \hline
\multicolumn{1}{r|}{}                           & \cellcolor[HTML]{EFEFEF}{\color[HTML]{000000} 5\%} & \cellcolor[HTML]{EFEFEF}{\color[HTML]{000000} 7/30} & \cellcolor[HTML]{EFEFEF}{\color[HTML]{000000} 3/30} & \cellcolor[HTML]{EFEFEF}{\color[HTML]{000000} 2/30} \\ \cline{2-5} 
\multicolumn{1}{r|}{\multirow{-1.9}{5em}{Nguyen-6}} & 10\%                                               & 10/30                                               & 4/30                                                & 3/30                                                \\ \hline
\multicolumn{1}{l|}{}                           & \cellcolor[HTML]{EFEFEF}5\%                        & \cellcolor[HTML]{EFEFEF}5/30                        & \cellcolor[HTML]{EFEFEF}4/30                        & \cellcolor[HTML]{EFEFEF}2/30                        \\ \cline{2-5} 
\multicolumn{1}{l|}{\multirow{-1.9}{4em}{Pagie-1}}  & 10\%                                               & 11/30*                                               & 9/30*                                                & 0/30                                                 \\ \hline
\end{tabular}
\end{adjustwidth}
\end{table}


\section{Discussion and additional remarks}

Within the limits of the current study, the results presented throughout the last section allow us to draw some conclusions that should be discussed in more detail. Although the central point of this work is not to establish a direct comparison between a Mate Choice mechanism (PIMP or Random Mate) and a Standard Approach, it is nonetheless relevant to make some notes on performance and diversity. It seems that in the long run (i.e., after 1500 generations), reducing fitness selective pressure does not affect significantly the performance, whether it is achieved by choosing a partner via preferences or at random. Comparing our results to that of the original work on PIMP, it seems that elongating the running time improves performance by a quite small margin in some symbolic regression instances. Arguably, mutation might also have something to do with this. As the original work provides a wider range of testing instances, it might be useful to extend the current experiment in the future. \par
Comparing PIMP directly against a random mate choice, the former seems to provide slightly better results regarding MBF, but we found these differences to be insufficient to reach statistical significance. Again,  testing a more diverse set of symbolic regression instances could be beneficial in this regard. As such, under the conditions applied in this experiment, it seems that PIMP and Random Mate are similar performance-wise. \par
On the other hand, in general, results regarding diversity favour PIMP. We believe this to be a particularly interesting -- and arguably counterintuitive -- finding: the dynamic evolution of preferences in PIMP seems to have a higher potential for promoting unique solutions rather than a simple random mate choice. This is also a topic to be addressed with more detail in the future, that is, a more focused analysis on how the evolvable preferences tend to promote more diversity than a random approach. Arguably, this might be a result of the directional Sexual Selection force that emerges throughout the evolutionary process, which in turn might be doing a better job at racing against fitness pressure than a non-directional random selection. \par
\begin{figure}[h]
  \centering
  \subfloat[PIMP]{\includegraphics[width=0.5\textwidth]{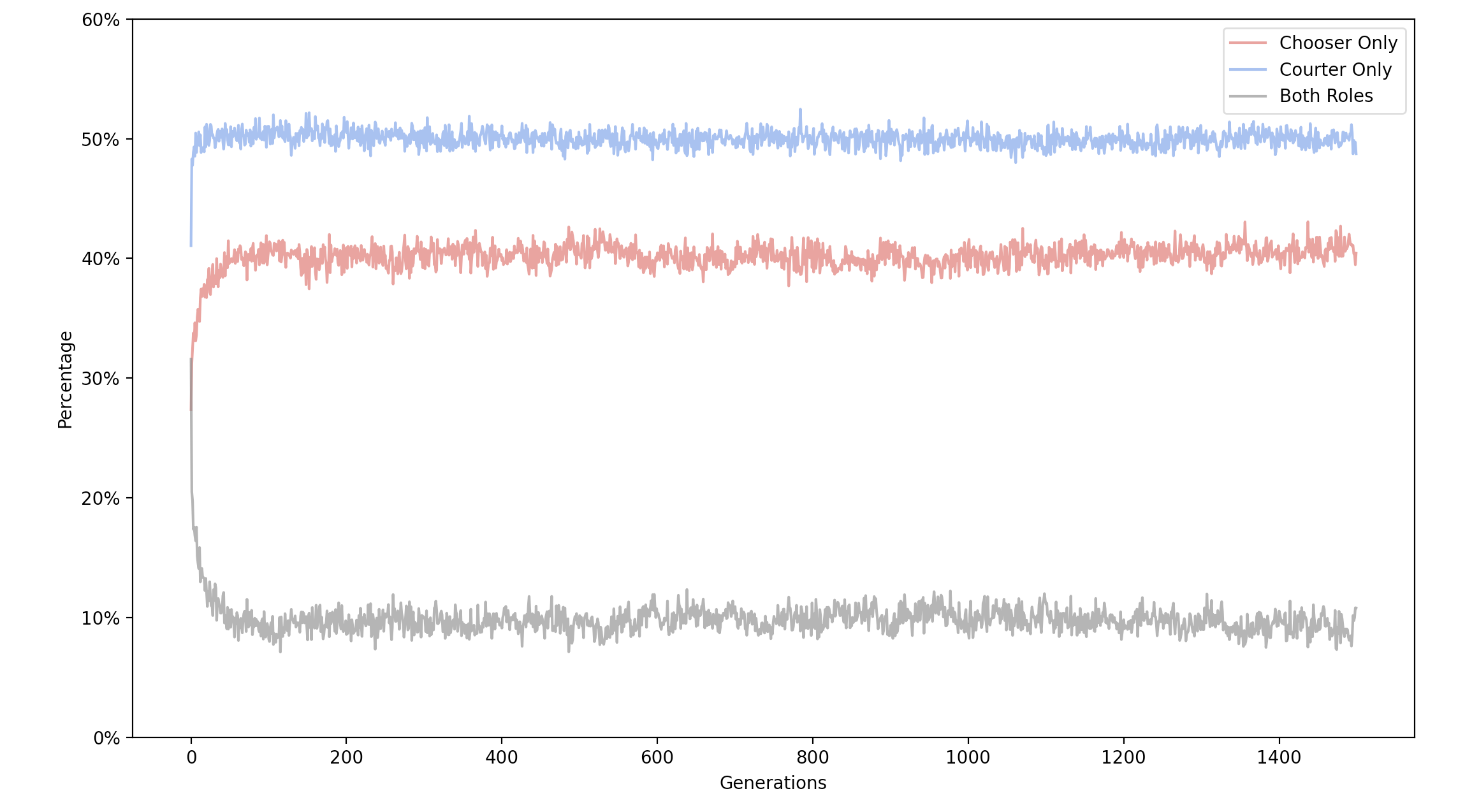}\label{pimp_gend_role}}
  \hfill
  \subfloat[Random Mate]{\includegraphics[width=0.48\textwidth]{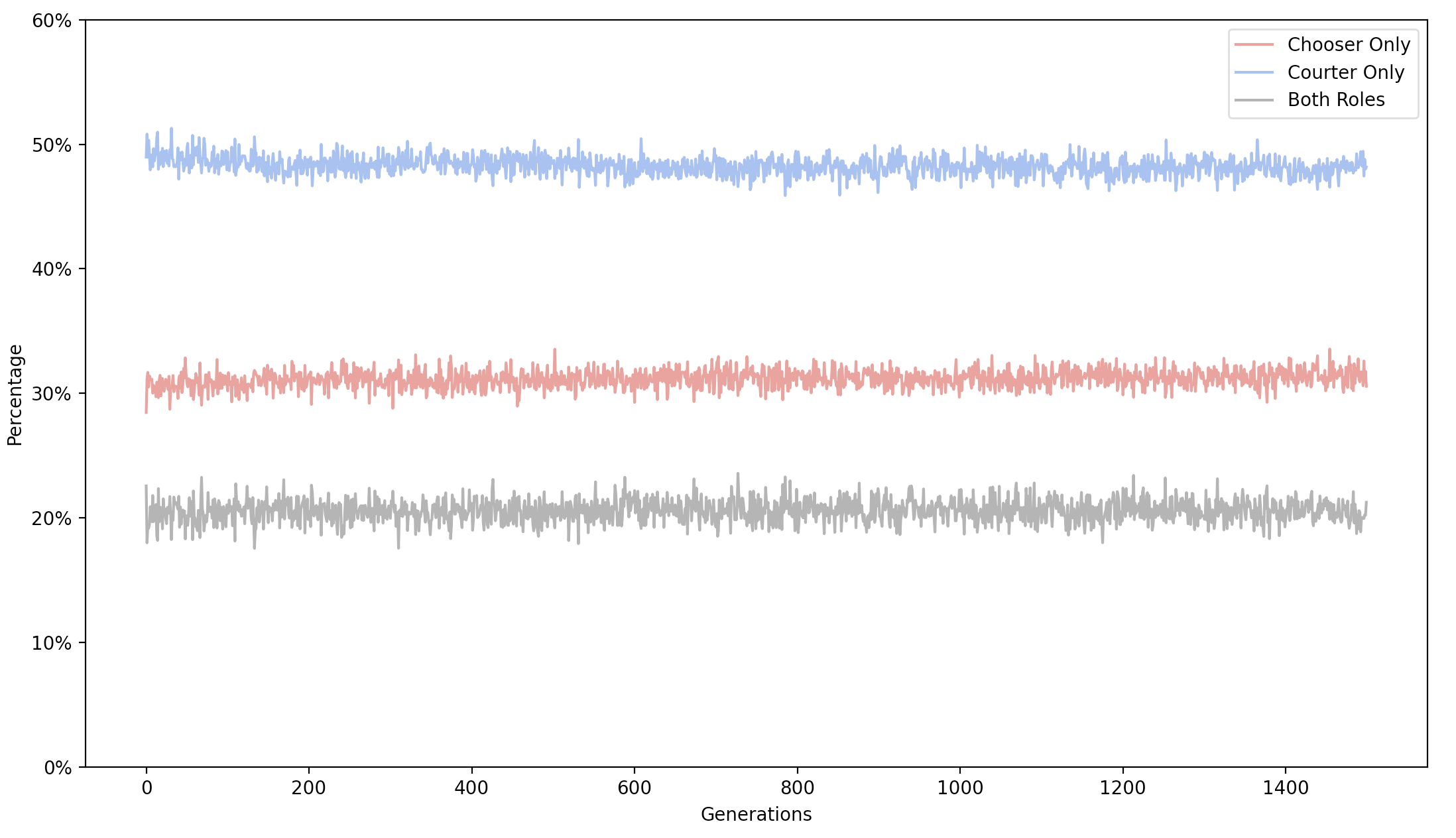}\label{random_gender_role}}
  \caption{Role Seggregation - Koza-1 Mutation rate 5\%.}
\end{figure}

Although this analysis requires an in-depth review, an answer to this difference may be laying precisely in how Sexual Selection via preferences shapes the search space of the ideal mates. In the original work of PIMP, a clear role separation throughout the evolutionary process was observed: the population tended to evolve into three main types (choosers, courters and both). To understand whether the same effect could be achieved via a Random Mate selection, we kept track of the selected individuals to establish a direct comparison. As such, we used the same terminology: Choosers (individuals that were only selected via tournament), Courters (individuals that were exclusively selected by a Chooser) and Both (individuals that in the same generation acted as a Chooser and a Courter). The evolutionary lineages of these can be seen in Figure \ref{pimp_gend_role} for PIMP and in Figure \ref{random_gender_role} for Random Mate selection. Note that only one instance is shown (Koza-1 with a mutation rate of 5\%), but the behaviour was similar in all symbolic regression instances. The small shifts observed in the first generations using the PIMP approach suggest an adaptation rather than a predefined rule (as seen in Random Mate, where on average all roles emerge in the same range in which they terminate). It is interesting to note that although a Random Mate choice does not provide a specific direction regarding choice, it manages to produce role segregation anyway. Another difference between these two figures is that Random Mate seems to be producing more individuals that act as Both roles, while in PIMP there’s a larger percentage of Choosers. \par

\begin{figure}[ht]
  \centering
  \subfloat[PIMP]{\includegraphics[width=0.7\textwidth]{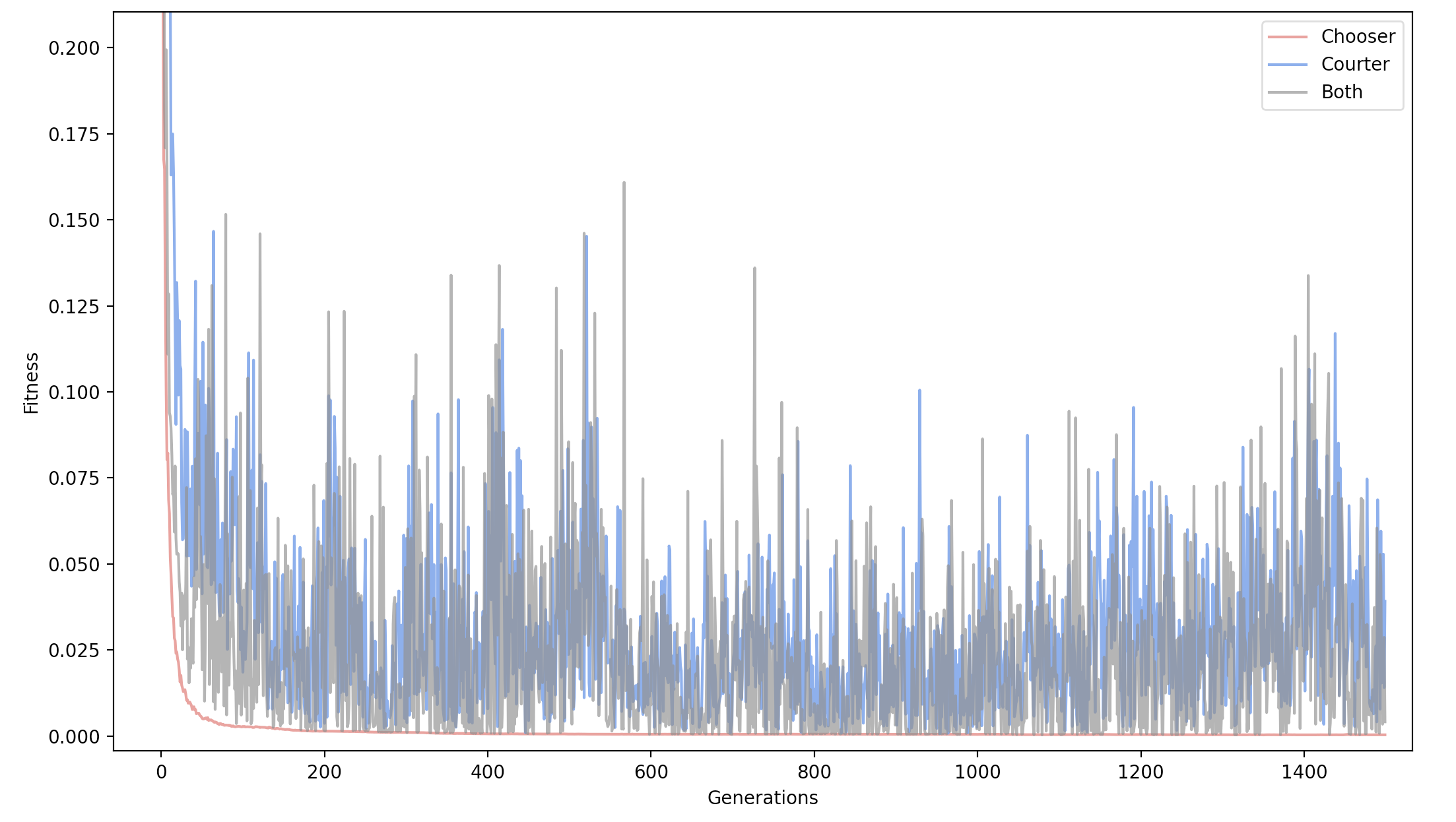}\label{pimp_mbf_sep}}
  \hfill
  \subfloat[Random Mate]{\includegraphics[width=0.7\textwidth]{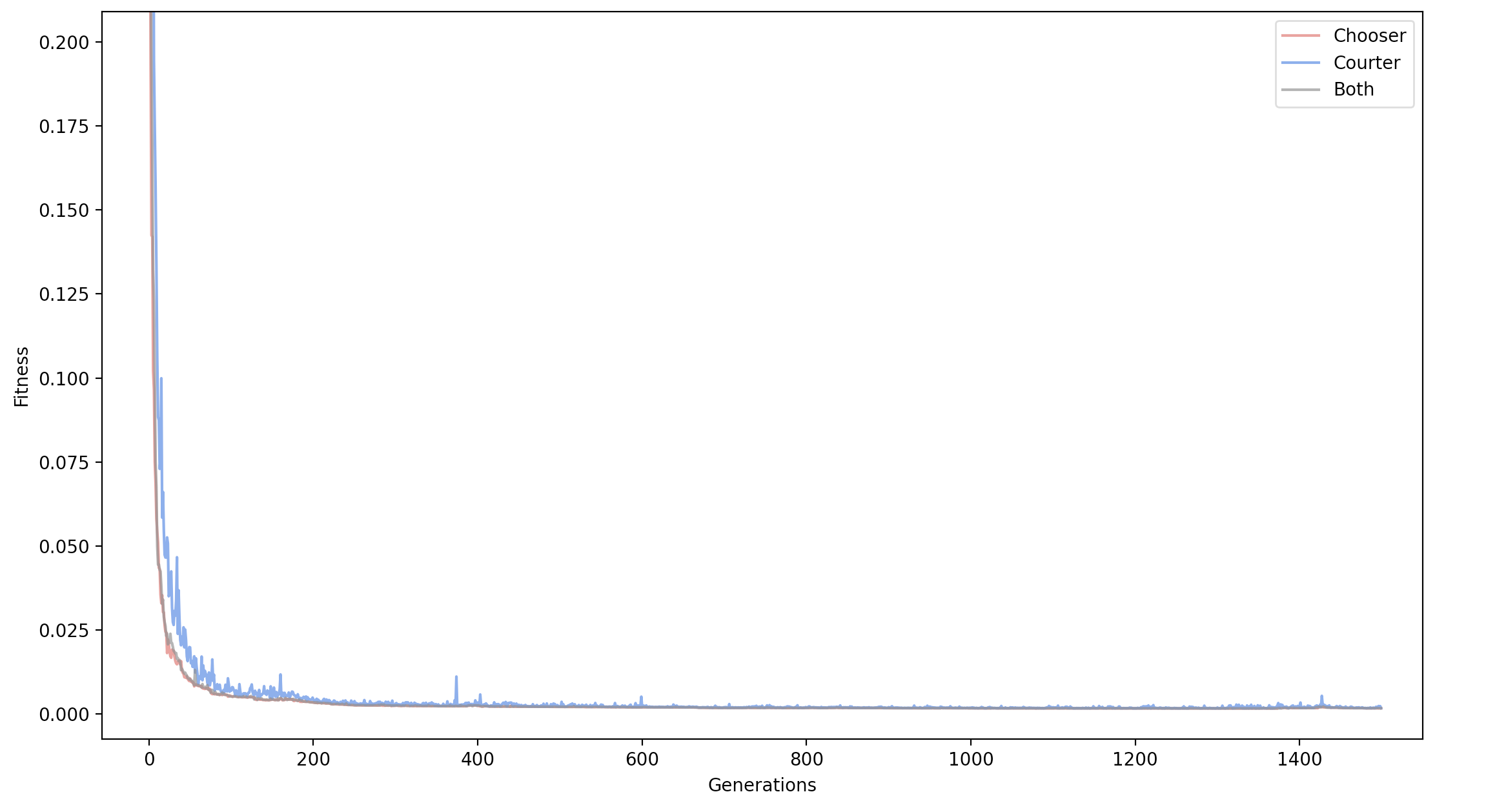}\label{rand_mbf_sep}}
  \caption{MBF evolution for different roles - Koza-1 Mutation rate 5\%.}
\end{figure}
We set out to analyse whether this stronger segregation had any impact on the MBF of each role (Figures \ref{pimp_mbf_sep}  and \ref{rand_mbf_sep}). \par
As expected, in both approaches the Choosers follow a better fitness path as they are subject to fitness pressure. Yet interestingly, a Random Mate choice seems more prone to preserve individuals with good fitness values while in PIMP these seem to be performing some sort of exploration. Furthermore, the MBF of individuals assuming both roles also seems to be evolving differently: In Random Mate, it looks like these behave as Choosers fitness-wise, while in PIMP this type of role assumes a more unpredictable behaviour, and these differences might be contributing to a more diverse population overall. By choosing a mate randomly, it means that the individuals that made it through the tournament have the same chance of being selected as a partner, while PIMP might be forcing a particular portion of the population to adapt to the existing range of chooser's preferences. Considering that even the chromosomes that encode the ideal partner are subject to mutations and crossover, this might also be contributing to the diversity of the whole population. As such, a detailed study on how PIMP shapes these separated roles might be valuable to understand the inner mechanisms of mating preferences in GP.

\section{Conclusion}

Diversity maintenance is known to be an important factor in EAs. Sexual Selection, particularly Mate Choice, has shown promising results when transposed to algorithmic practice, yet its dynamics and potential benefits are poorly understood, especially when it comes to GP. In this article, we took advantage of PIMP -- mate choice based on ideal mates -- to address whether there were tangible differences between guiding the evolutionary process via a random mate choice and by having a set of individual mating preferences. \par
After 1500 generations (in sets of 30 runs), PIMP revealed competitive results regarding MBF and had a better success rate in half of the testing instances. Although there were no statistically significant differences regarding performance, these were achieved when considering diversity gains (particularly in the average number of unique solutions), where PIMP has shown a better capacity of doing so at least in the long run. As such, we can conclude that within our experiments PIMP promoted more diversity within populations without compromising performance.
These results, coupled with other indicative factors such as root node diversity (although in this regard fewer differences were statistically significant), seem to suggest that there are in fact differences between PIMP and a Random Mate choice. Moreover, the recorded differences are interesting: a mate choice based on ideal mate preferences seems to be promoting more diversity within the population (even fitness-wise) than a random mate search. We believe these to be meaningful results in the field of Sexual Selection applied to GP, which deserves an in-depth study on how preferences evolve, mainly on the evolutionary dynamics of choosers and courters. By doing so, we may not only understand in more detail how this force is promoting more diversity than a random mate choice but also if (and how) this approach can be shaped to improve algorithmic performance.\par
Finally, it must be noted that for such an intricate task, it would be beneficial in the future to have a more broad study on the comparison of these two approaches, mainly with a broader set of symbolic regression instances.

\section*{}
\textbf{Acknowledgements.} This work is funded by national funds through the FCT - Foundation for Science and Technology, I.P., within the scope of the project CISUC - UI/BD/151046/2021 and by European Social Fund, through the Regional Operational Program Centro 2020



%
%
%
%

\end{document}